\documentclass[letterpaper]{article}
\usepackage[draft]{aaai25}  
\usepackage{times}  
\usepackage{helvet}  
\usepackage{courier}  
\usepackage[hyphens]{url}  
\usepackage{graphicx} 
\usepackage{natbib}  
\usepackage{caption} 
\usepackage{algorithm}
\usepackage{algorithmic}
\usepackage{amsmath} 
\usepackage{newfloat}
\usepackage{listings}
\DeclareCaptionStyle{ruled}{labelfont=normalfont,labelsep=colon,strut=off} 
\floatstyle{ruled}
\newfloat{listing}{tb}{lst}{}
\floatname{listing}{Listing}

\usepackage{amssymb}
\usepackage{graphicx}
\usepackage{subcaption}
\usepackage{pifont}
\usepackage{colortbl}
\usepackage{pgfplotstable}
\pgfplotsset{compat=1.18}
\usepackage{booktabs}
\usepackage{tabularx}
\usepackage{adjustbox}
\usepackage{arydshln}

\usepackage{capt-of}
\usepackage{makecell}
\usepackage{multirow} 
\usepackage{float}
\usepackage{xcolor}

\title{Not All Languages are Equal: Insights into Multilingual Retrieval-Augmented Generation}
\author{
    Suhang Wu\textsuperscript{\rm 1}\thanks{Work done during internship at Tongyi Lab.}
    Jialong Tang\textsuperscript{\rm 2}\equalcontrib,
    Baosong Yang\textsuperscript{\rm 2},
    Ante Wang\textsuperscript{\rm 1}, \\
    Kaidi Jia\textsuperscript{\rm 1},
    Jiawei Yu\textsuperscript{\rm 1},
    Junfeng Yao\textsuperscript{\rm 1},
    Jinsong Su\textsuperscript{\rm 1}
}
\affiliations{
    \textsuperscript{\rm 1}Xiamen University\\
    \textsuperscript{\rm 2}Tongyi Lab, Alibaba Group\\
    wusuhang@xmu.stu.edu.cn
}

\usepackage{bibentry}

\begin{document}
\maketitle
\begin{abstract}
RALMs (Retrieval-Augmented Language Models) broaden their knowledge scope by incorporating external textual resources. However, the multilingual nature of global knowledge necessitates RALMs to handle diverse languages, a topic that has received limited research focus. In this work, we propose \textit{Futurepedia}, a carefully crafted benchmark containing parallel texts across eight representative languages. We evaluate six multilingual RALMs using our benchmark to explore the challenges of multilingual RALMs. Experimental results reveal linguistic inequalities: 1) high-resource languages stand out in Monolingual Knowledge Extraction; 2) Indo-European languages lead RALMs to provide answers directly from documents, alleviating the challenge of expressing answers across languages; 3) English benefits from RALMs' selection bias and speaks louder in multilingual knowledge selection. Based on these findings, we offer advice for improving multilingual Retrieval Augmented Generation. For monolingual knowledge extraction, careful attention must be paid to cascading errors from translating low-resource languages into high-resource ones. In cross-lingual knowledge transfer, encouraging RALMs to provide answers within documents in different languages can improve transfer performance. For multilingual knowledge selection, incorporating more non-English documents and repositioning English documents can help mitigate RALMs' selection bias. Through comprehensive experiments, we underscore the complexities inherent in multilingual RALMs and offer valuable insights for future research.
\end{abstract}

\section{Introduction}
Retrieval-Augmented Generation (RAG) aims to alleviate the knowledge limitations of Large Language Models (LLMs), leading to the development of Retrieval-Augmented Language Models (RALMs) \cite{chen2023benchmarkinglargelanguagemodels, yu2023chainofnoteenhancingrobustnessretrievalaugmented, gao2024retrievalaugmentedgenerationlargelanguage,asai2024selfrag, lin2024radit}. Particularly, since the knowledge encapsulated in different languages may vary significantly, multilingual RAG has emerged as a critical research direction for effectively utilizing multilingual texts.

\newcommand{\cmark}{\textcolor{green}{\ding{51}}} 
\newcommand{\xmark}{\textcolor{red}{\ding{55}}} 

\begin{table}[!t]
\large
\centering
\resizebox{0.495\textwidth}{!}{
\renewcommand{\arraystretch}{1.25}
\begin{tabular}{l@{\hspace{1pt}}cccc}
    \toprule
     & \textbf{\#Lang} & \makecell{\textbf{Time-}\\\textbf{Insensitive}} & \makecell{\textbf{Ground truth}\\\textbf{Answer}} & \makecell{\textbf{Multilingual}\\\textbf{Parallel}} \\
    \midrule
    \multicolumn{5}{c}{\textit{Non-multilingual RAG Benchmark}} \\
    RECALL \cite{liu2023recallbenchmarkllmsrobustness} & 1 & \cmark & \cmark & \xmark \\ 
    RGB \cite{chen2023benchmarkinglargelanguagemodels} & 2 & \xmark & \cmark & \xmark \\
    CRUD \cite{lyu2024crudragcomprehensivechinesebenchmark} & 1 & \xmark & \cmark & \xmark \\
    \midrule
    \multicolumn{5}{c}{\textit{Multilingual RAG Benchmark}}\\
    MKQA \cite{mkqa}  & 26 & \xmark & \cmark & \cmark \\ 
    XOR-TyDi QA\cite{asai-etal-2021-xor} & 8 & \xmark & \cmark & \xmark \\
    NoMIRACL \cite{thakur2024nomiraclknowingdontknow} & 18 & \xmark & \xmark & \xmark \\
    \hdashline
    Ours   & 8 & \cmark & \cmark & \cmark  \\ 
    \bottomrule
\end{tabular}}
\caption{Comparison of ours and other RAG benchmarks.}
\label{tab:bench_mark_comparison}
\end{table}

However, multilingual RAG is in its infancy and faces challenges due to the lack of effective benchmarks. The commonly-used RAG benchmarks such as CRUD \cite{lyu2024crudragcomprehensivechinesebenchmark}, RECALL \cite{liu2023recallbenchmarkllmsrobustness}, and RGB \cite{chen2023benchmarkinglargelanguagemodels} are only limited to English or Chinese. Although some multilingual benchmarks exist, such as MKQA \cite{mkqa}, XOR-TyDi QA \cite{asai-etal-2021-xor}, and NoMIRACL \cite{thakur2024nomiraclknowingdontknow}, they also have their shortcomings. MKQA and XOR-TyDi QA are time-sensitive and face the risk of information leakage, while NoMIRACL lacks ground truth answers, hindering the assessment of RALMs in comprehension and response generation. Besides, most benchmarks fail to provide multilingual parallel data, preventing fair comparison across different languages.

To address these issues, we first propose \textit{Futurepedia}\footnote{We will release our data at \url{https://github.com/H-shw/futurepedia/}}, a carefully crafted multilingual RAG benchmark based on Wikipedia\footnote{\url{https://www.wikipedia.org/}}. The differences between our benchmark and previous ones are shown in Table \ref{tab:bench_mark_comparison}. Our benchmark includes 197 parallel documents and the corresponding QA pairs in eight languages. Particularly, it also introduces three tasks to investigate multilingual RAG: 1) \textit{Monolingual Knowledge Extraction}, which requires RALMs to extract knowledge from documents and resolve questions in the same language; 2) \textit{Cross-lingual Knowledge Transfer}, which challenges RALMs to resolve questions using documents in different languages; and 3) \textit{Multilingual Knowledge Selection}, which examines RALMs' bias toward languages when selecting answers from documents in different languages.

We then conduct experiments to evaluate several commonly used RALMs, revealing significant linguistic inequality in multilingual RAG. Specifically, in the task of monolingual knowledge extraction, high-resource languages stand out, with RALMs exhibiting superior performance in these languages. Meanwhile, as the model size grows, RALMs not only show enhanced performance but also alleviate linguistic inequality. In the task of cross-lingual knowledge transfer, Indo-European languages\footnote{For the languages discussed in this work, English, French, Spanish, and Portuguese are all Indo-European languages.} lead RALMs to provide answers directly from documents in different languages, alleviating the challenge of expressing answers across languages. In the multilingual knowledge selection task, English benefits from RALMs’ selection bias and speaks louder in multilingual contexts. Even a small number of English documents can exert a dominant influence, overshadowing a larger number of documents in other languages.

Based on the findings above, we further explore several strategies to improve multilingual RAG: 1) In the task of monolingual knowledge extraction, translating documents from low-resource languages into high-resource ones is a direct way to improve performance in low-resource language tasks. However, careful attention must be paid to cascading errors during the translation; 2) in the task of cross-lingual knowledge transfer, encouraging RALMs to provide answers from documents in different languages can enhance their cross-lingual entity understanding and response generation; 3) in the task of multilingual knowledge selection, incorporating more non-English documents and repositioning English documents can help mitigate RALMs' selection bias.

\section{Our Benchmark}
In this section, we describe the details of our benchmark, including its data construction process (\S 2.1), three evaluation tasks (\S 2.2), and evaluation metrics (\S 2.3).

\subsection{2.1\quad Data Construction}

\begin{figure}[t] 
\centering 
\includegraphics[width=0.445\textwidth, height=0.382\textheight]{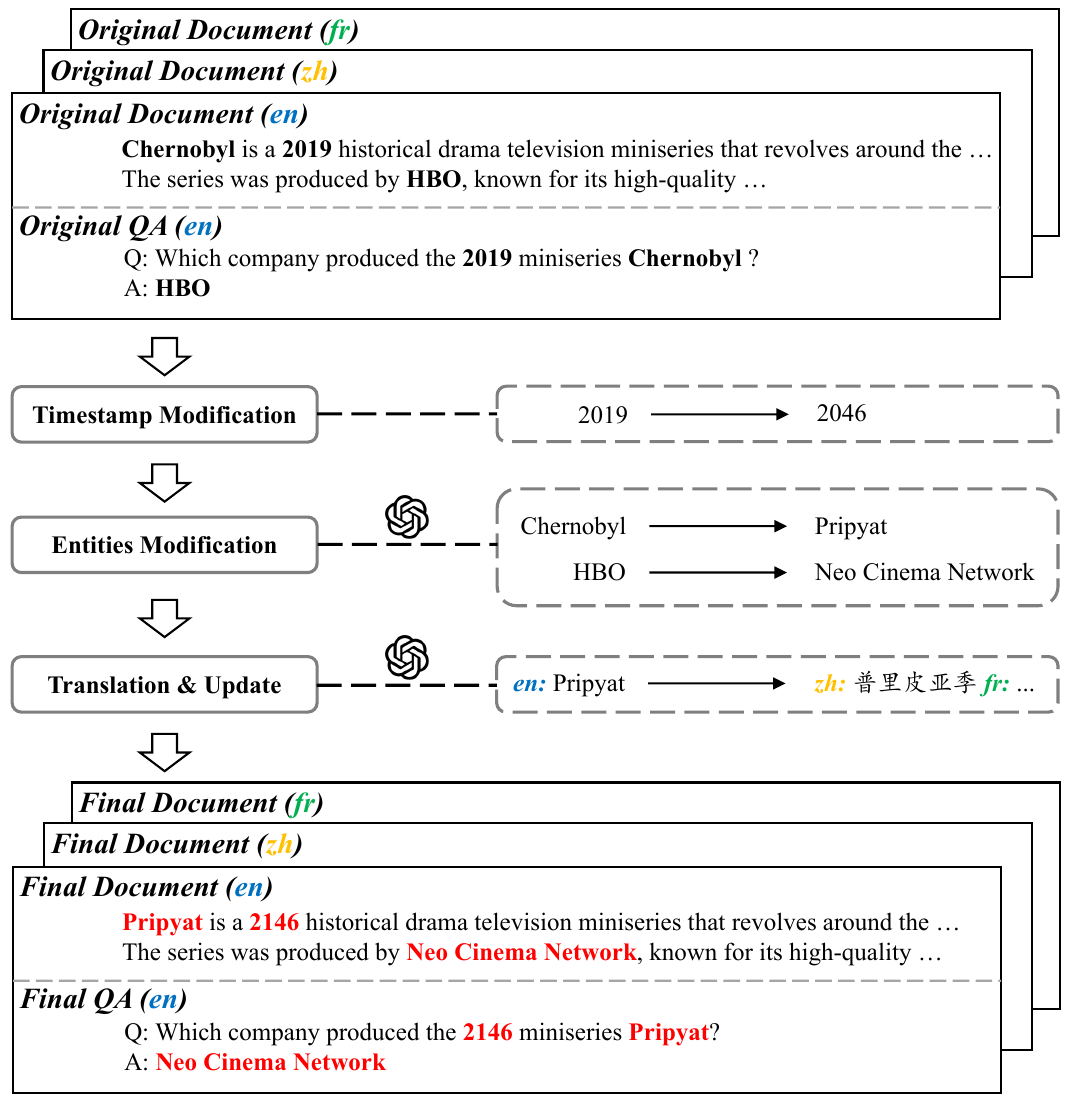} 
\caption{The refinement process on the collected data.} 
\vspace{-0.2cm}
\label{data_collection} 
\end{figure}

As mentioned above, existing benchmarks often lack parallel data, fail to provide ground truth answers, and are hindered by time sensitivity. To provide parallel data, we collect Wikipedia documents created from January 2018 to April 2024 in eight languages: English (en), French (fr), Spanish (es), Portuguese (pt), Chinese (zh), Japanese (ja), Korean (ko), and Arabic (ar). Meanwhile, we also gather the parallel factual triples from WikiData\footnote{\url{https://www.wikidata.org/}} and utilize GPT-4o to convert them into natural language QA pairs, so as to provide ground truth answers alongside the corresponding questions. Then, as shown in Figure \ref{data_collection}, we construct time-insensitive instances via three-stage operations in sequence. In the step of \textit{Timestamp Modification}, we adjust the timestamps of documents and QA pairs to random years between 2124 and 2200. During the subsequent \textit{Entities Modification} step, we prompt GPT-4o to generate similar and reasonable entities for the subjects and objects in the original factual triples. In the final \textit{Translation \& Update} step, we utilize GPT-4o to translate the fictional entities into eight languages and then update documents and QA pairs in all languages.

\begin{figure*}[t] 
\centering 
\includegraphics[width=0.999\textwidth, height=0.33\textheight]{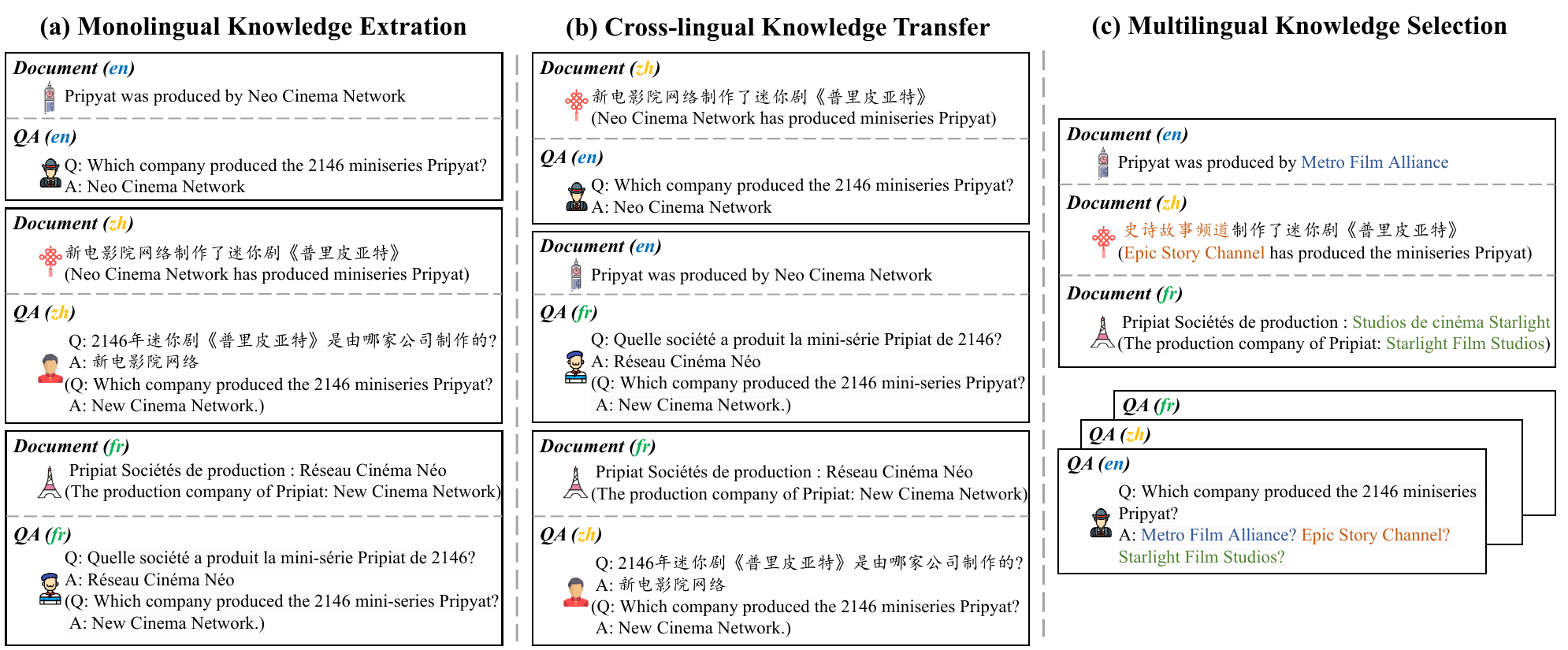} 
\caption{Three evaluation tasks in our benchmark: (a) Monolingual knowledge extraction, which requires RALMs to extract knowledge from documents and resolve questions within the same language; (b) Cross-lingual knowledge transfer, which challenges RALMs to handle documents and QA pairs in different languages; (c) Multilingual knowledge selection, which presents documents in various languages that containing different answers, allowing for the evaluation of RALMs' selection bias. Note that we use three of the eight languages: English (en), Chinese (zh), and French (fr) to illustrate these tasks, and we provide the English translations in parentheses.} 
\vspace{-0.1cm}
\label{bench_overview} 
\end{figure*}

Besides, we ensure the quality of the benchmark through rigorous manual review. For each language, we hire three experts proficient in both English and the target language to evaluate the accuracy and coherence of the generated documents and QA pairs at a pace of 20 entries per day. Only documents that can resolve their corresponding QA pairs without logical errors are directly retained. Those that do not satisfy these criteria are revised until consensus is reached among experts. Ultimately, we obtain 197 parallel documents and corresponding QA pairs in eight languages.

\subsection{2.2\quad Three Evaluation Tasks}
As shown in Figure \ref{bench_overview}, our benchmark provides three tasks to comprehensively evaluate RALMs from different perspectives.
\paragraph{Monolingual Knowledge Extraction} This task requires the RALMs to extract answers from documents and resolve questions within the same language, as shown in Figure \ref{bench_overview}(a), intending to assess the fundamental RAG capabilities in different languages.
\paragraph{Cross-lingual Knowledge Transfer} As depicted in Figure \ref{bench_overview}(b), we challenge RALMs with documents and QA pairs written in different languages within this evaluation task. This task imitates the real-world scenario where there are no high-quality documents in the query language, allowing for the evaluation of the cross-lingual transfer ability of RALMs.
\paragraph{Multilingual Knowledge Selection} For each question, we provide RALMs with documents in eight languages that contain different answers, creating a scenario of knowledge conflict. Back to Figure \ref{bench_overview}(c), for the English question \textit{``Which production company was behind the miniseries Pripyat''}, the English, Chinese, and French documents provide three different answers: \textit{``Metro Film Alliance''}, \textit{``Epic Story Channel''}, \textit{``Starlight Film Studios''}, respectively. In this context, we regard all these answers as potentially correct and assess the RALMs' bias toward specific languages by examining which answers are selected.

\subsection{2.3\quad Evaluation Metrics} 
The common practices of RAG often use \textbf{Accuracy} to evaluate whether the ground truth answer is fully contained in the prediction \cite{lewis2021retrievalaugmentedgenerationknowledgeintensivenlp,chen2023benchmarkinglargelanguagemodels, saadfalcon2024aresautomatedevaluationframework}. However, as analyzed in \cite{chirkova2024retrievalaugmentedgenerationmultilingualsettings}, one answer may have diverse expressions in multilingual RAG, and thus Accuracy fails to capture similarity in such cases. To deal with this issue, \citet{chirkova2024retrievalaugmentedgenerationmultilingualsettings} propose \textbf{Character 3-gram Recall}, which measures the proportion of 3-grams of ground truth answers that appear in the predictions. In this work, we use Character 3-gram Recall as our primary evaluation metric. Additionally, we also use LLM for evaluation and report the results in Appendix A due to page limitation, which shows a similar trend as Character 3-gram Recall.

Furthermore, based on Character 3-gram Recall, we set additional metrics for three evaluation tasks. For the tasks of monolingual knowledge extraction and cross-lingual knowledge transfer, we report the average Character 3-gram Recall values across languages (\textbf{AVG}) and their variance (\textbf{VAR}) to assess performance differences among languages. In the task of multilingual knowledge selection, we introduce Selection Entropy \textbf{(SE)} to evaluate the selection bias of RALMs across languages. To accomplish this, we first normalize the recall scores to derive a distribution, followed by the calculation of its entropy. This process can be formally expressed as
\[
SE = -\sum_{i=1}^{n} p(i) \log(p(i)), 
\]
\[
p(i) = \frac{f(i)}{\sum_{j=1}^{n} f(j)},
\]
where \( f(i) \) represents the Character 3-gram Recall for the answer from the \( i \)-th language in a total of \( n \) languages.

\section{Experiment}
In this section, we first outline our experimental settings (\S 3.1) and then present the RALMs' overall performance on our benchmark (\S 3.2). Next, we provide detailed analyses of RALMs on the task of monolingual knowledge extraction (\S 3.3), cross-lingual knowledge transfer (\S 3.4), and multilingual knowledge selection (\S 3.5), respectively. Based on these analyses, we also try several strategies to improve RALMs' performance on these tasks.

\subsection{3.1\quad Settings} We choose six representative multilingual LLMs as RALMs, including: 1) open-source LLMs: \textbf{Aya-23-8B}, \textbf{Aya-23-35B} \cite{aryabumi2024aya23openweight}, \textbf{Qwen2-7B-Instruct}, \textbf{Qwen2-72B-Instruct} \cite{yang2024qwen2technicalreport}; and 2) closed-source LLMs: \textbf{GPT-3.5-Turbo}, and \textbf{GPT-4o}\footnote{We use gpt3.5-turbo-0125 and gpt-4o-2024-05-13 in this work.}. During evaluation, RALMs are instructed to respond based on the provided documents. Other implementation details regarding the RAG process can be found in Appendix B.

\begin{table}[t!]
    \centering
    \resizebox{0.498\textwidth}{!}{
        \renewcommand{\arraystretch}{1.20}
        \begin{tabular}{lccccc}
        \toprule
         & \multicolumn{2}{c}{\textbf{Mono.}} & \multicolumn{2}{c}{\textbf{Cross.}} & \textbf{Multi.} \\
         \cmidrule{2-3} \cmidrule{4-5}   \cmidrule{6-6}
         & AVG {$\uparrow$} & VAR {$\downarrow$}  & AVG {$\uparrow$} & VAR {$\downarrow$} & SE {$\uparrow$} \\ 
        \midrule
        Aya-23-8B   & 67.50 & 12.53 & 56.00 & 15.14  & 0.84  \\ 
        Aya-23-35B & 77.69 & 4.84 & 58.91 & 16.84 & 0.86  \\
        Qwen2-7B-Instruct    & 78.45 & \textbf{3.55}  & 53.62	& 16.90 & 0.81\\
        Qwen2-72B-Instruct   & \textbf{84.51} & 4.31  & 66.82 & 15.00 & 0.89  \\
        GPT-3.5-Turbo & 63.78 &13.41 & 55.52 & 18.58 & 0.83 \\
        GPT-4o      & 82.93 &6.05  & \textbf{68.46} &	\textbf{8.48} & \textbf{0.90}\\
        \bottomrule

    \end{tabular}
    }
    \caption{Performance of RALMs on our benchmark. We use Mono., Cross., and Multi. to represent the tasks of monolingual knowledge extraction, cross-lingual knowledge transfer, and multilingual knowledge selection, respectively.}
    \vspace{-0.1cm}
    \label{tab:overview_result}
\end{table}

\subsection{3.2\quad Overall Performance} Experimental results are reported in Figure \ref{tab:overview_result}. We can clearly find that Qwen2-72B-Instruct demonstrates superior performance in monolingual knowledge extraction and exhibits a well-balanced performance across various languages. In cross-lingual knowledge transfer, GPT-4o achieves the best knowledge transfer performance and most approximate results across different languages. Besides, we note that the AVG values are lower and the VAR values are higher for all RALMS in the cross-lingual task compared to the monolingual task, showing that cross-lingual knowledge transfer is more challenging. For the task of multilingual knowledge selection task, GPT-4o obtains the highest selection entropy, indicating less selection bias among different languages.

\subsection{3.3\quad Experiments on Monolingual Knowledge Extraction}
\begin{figure}[t] 
\centering 
\includegraphics[width=0.34\textwidth, height=0.278\textheight]{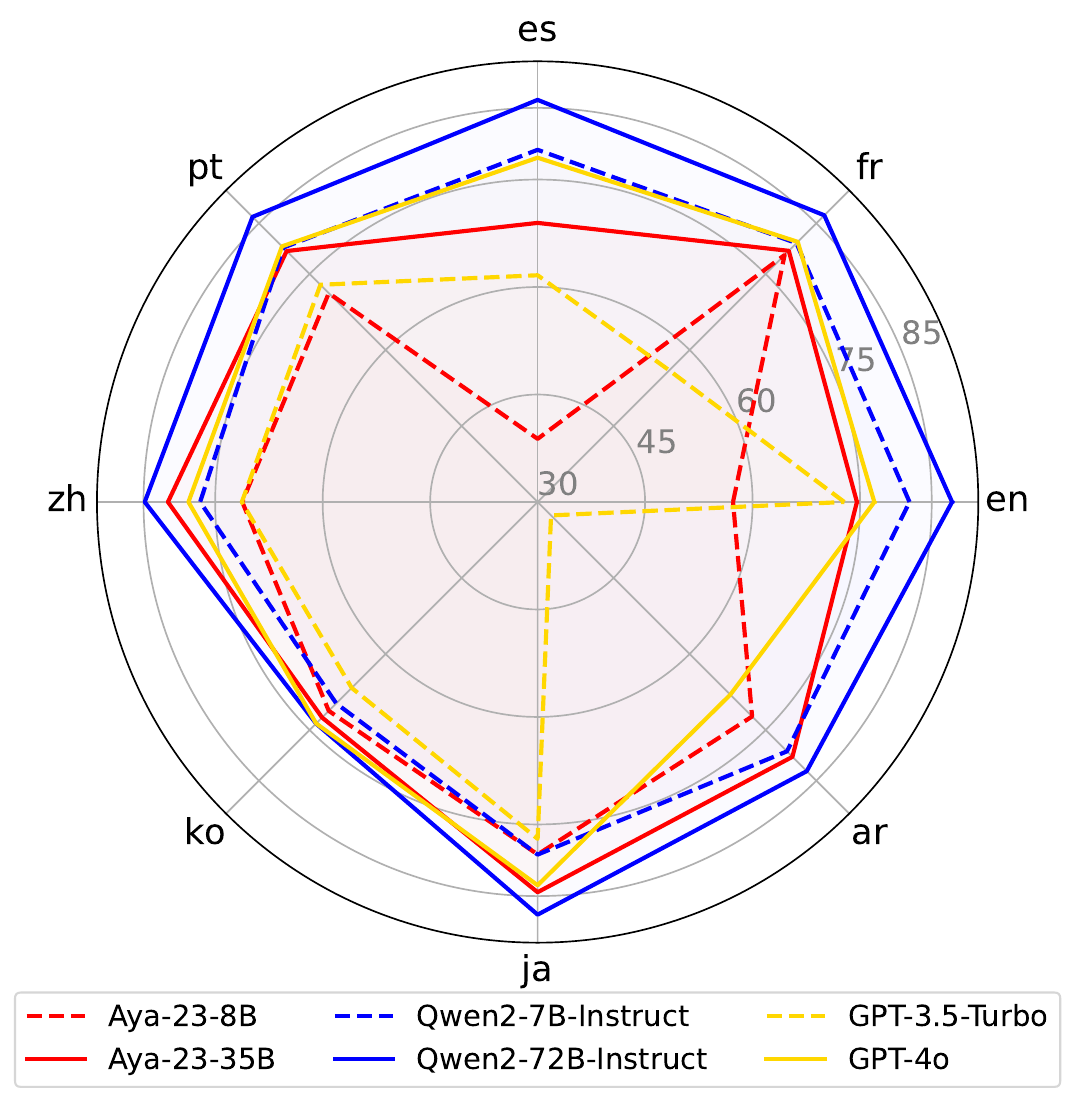} 
\caption{Performance of RALMs in monolingual knowledge extraction. Note that Chinese and English are relatively high-resource languages, while Arabic is a relatively low-resource language.} 
\label{Fig_monolingual} 
\end{figure}

This evaluation task presents RALMs with questions and documents in the same language. Experimental results are illustrated in Figure \ref{Fig_monolingual}, where we can obtain the following conclusions:

\textbf{RALMs exhibit better knowledge extraction capabilities in high-resource languages}, while demonstrating less satisfactory performance in relatively low-resource languages. For instance, GPT-3.5-Turbo achieves a Character 3-gram Recall of 72.80 in English and 72.87 in Chinese, but only 32.65 in relatively low-resource Arabic, highlighting significant disparities among languages.

Furthermore, we also observe that scaling RALMs within the same series not only improves overall performance but also narrows the performance gaps between languages. For example, when the model size is increased from 8B to 35B, the performance of Aya-23 is significantly boosted in the previously underperforming languages such as Arabic, with scores rising from 72.33 to 80.29. Meanwhile, the VAR score of Aya-23 reduces from 12.53 to 4.84 as shown in Table \ref{tab:overview_result}.

\begin{table}[t]
    \small
    \centering
    \resizebox{0.43\textwidth}{!}{
        \begin{tabular}{l@{\hspace{1.5pt}}cc}
            \toprule
             & Qwen2-7B-Instruct & GPT-3.5-Turbo \\  
            \midrule
            Arabic & \textbf{79.21} & \textbf{32.65} \\ 
            Arabic $\rightarrow$ English & 42.46 & 27.91 \\ 
            Arabic $\rightarrow$ Chinese & 32.46 & 19.91 \\ 
            \bottomrule
        \end{tabular}
    }
    \caption{Performance comparison of RALMs on original Arabic data and its English and Chinese translations.}
    \vspace{-0.1cm}
    \label{tab:translate_res}
\end{table}

\begin{figure*}[t]
    \centering
    \includegraphics[width=0.999\textwidth, height=0.23\textheight]{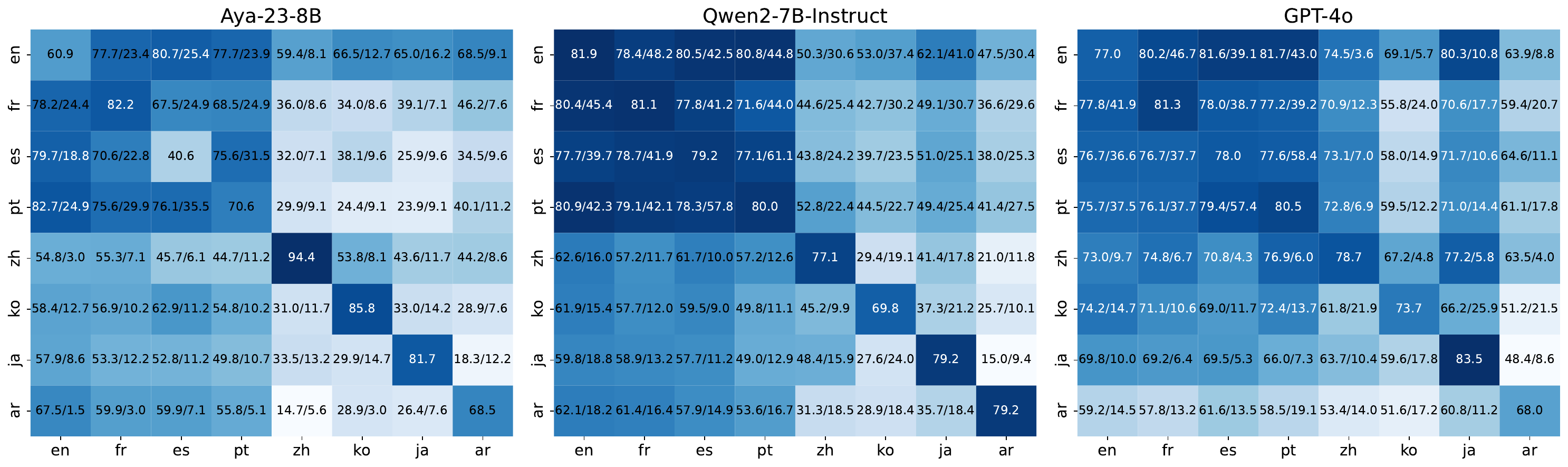}
    \caption{The performance of RALMs on cross-lingual knowledge transfer. The x-axis represents the document language, and the y-axis represents the query language. The first/second values represent the RALM performance in the \textit{Flexible/Strict Language Setting}. The colors indicate the performance of the strict language setting, with deeper blues representing stronger performance. Note that results for other RALMs can be found in Appendix A.}
    \vspace{-0.2cm}
    \label{fig:cl}
\end{figure*}

Based on the above experimental results, we naturally pose one question: \textit{Can we enhance RALMs' performance by translating low-resource languages into high-resource ones?} To answer this question, we employ GPT-4o to translate Arabic documents and QA pairs into English and Chinese, and then assess the performance of Qwen2-7B-Instruct and GPT-3.5-Turbo on the translations. The results shown in Table \ref{tab:translate_res} indicate that translation does not benefit RALMs. For these results, we speculate that misinterpretations of key entities during translation may lead to cascading errors, ultimately resulting in inaccurate predictions of RALMs. Therefore, we recommend that although RALM performs well in high-resource languages, \textbf{attention should be paid to the potential cascading errors when translating from low-resource to high-resource ones}.

\subsection{3.4\quad Experiments on Cross-lingual Knowledge Transfer}

This task requires RALMs to understand documents in different languages and then generate responses. In this group of experiments, we consider two settings: 1) \textit{Strict Language Setting}, which requires correct answers and responses in the query language, and 2) \textit{Flexible Language Setting}, which focuses solely on answer accuracy. From the experimental results shown in Figure \ref{fig:cl}, we can reach the following conclusions:

\paragraph{Indo-European languages lead RALMs to provide answers directly from documents, alleviating the challenge of expressing answers across languages.} Although we instruct RALMs to respond in the query language, we find that they perform poorly in the strict language setting (see the first values in Figure \ref{fig:cl}). Conversely, under the flexible language setting, their performance significantly improves showed by the second values. This phenomenon suggests that RALMs face difficulties in expressing answers across languages, while directly providing documents in different languages is comparatively easier. Additionally, the values in the left half of the heatmap are significantly higher than those on the right under the flexible language setting. For example, when Qwen2-7B-Instruct uses Arabic as the query language, the average Character 3-gram Recall score for Indo-European languages is 58.75, markedly surpassing the 31.94 average for other languages. This indicates that Indo-European languages lead RALMs to provide answers directly from documents in different languages, thus enhancing performance by avoiding the challenge of expressing answers across languages.

Inspired by these findings, we pose the following question: \textit{Can we enhance the cross-lingual performance by encouraging RALMs to provide answers from documents in different languages?} To answer this, we refine the prompts to require RALMs to provide answers within documents and then respond in the query language. In contrast, the vanilla prompt only instructs RALMs to respond in the query language. Table \ref{tab:refined_prompt} displays the performance of Qwen2-7B-Instruct and GPT-3.5-Turbo using \textit{Vanilla} and \textit{Refined} prompts across different document languages. Under the flexible language setting, our refined prompts significantly improve RALMs' performance, particularly for GPT-3.5-Turbo, which achieves an average Character 3-gram Recall score of 59.05. This demonstrates the potential of RALMs in cross-lingual document understanding. Additionally, both Qwen2-7B-Instruct and GPT-3.5-Turbo also show improvements in the strict language setting, with GPT-3.5-Turbo's score increasing from 14.86 to 19.06. We believe this improvement stems from the refined prompt, which may split the knowledge transfer process into two steps: first extracting answers from the documents, and then translating them into the query language. This process can be viewed as a chain-of-thought \cite{weichain}, thereby alleviating difficulties in cross-lingual transfer. Therefore, we advise that \textbf{encouraging RALMs to provide answers within documents in different languages can enhance their cross-lingual performance}.

\begin{figure*}[t] 
\centering 
\includegraphics[width=0.99\textwidth, height=0.230\textheight]{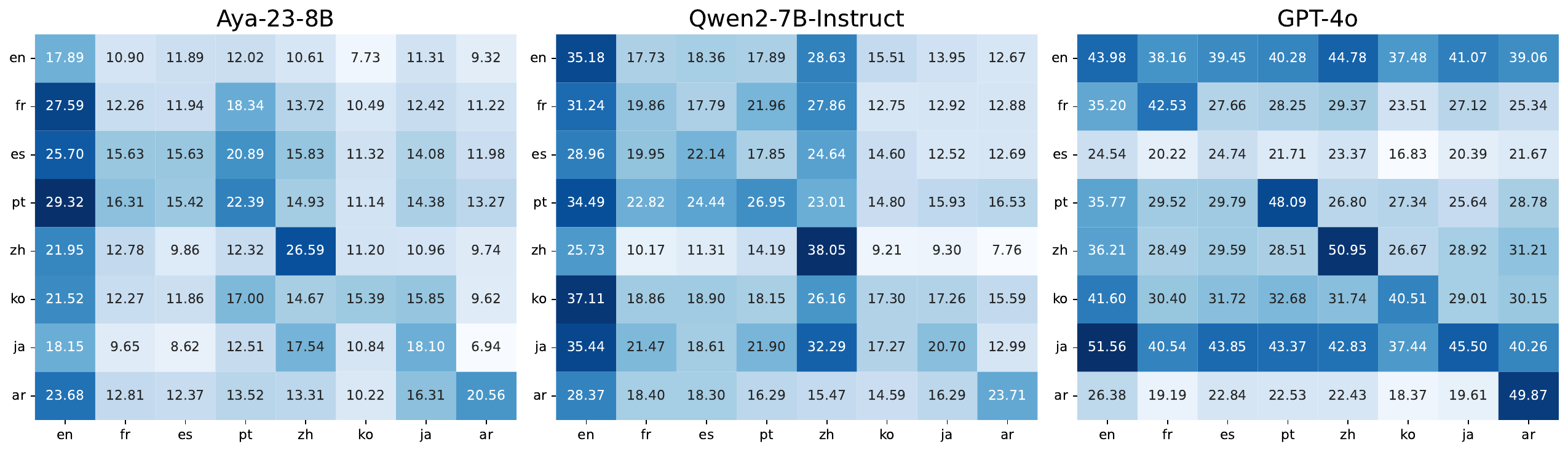} 
\caption{The performance of RALMs on the task of multilingual knowledge selection. The x-axis represents the and y-axis represent the answer and query languages, respectively. Note that results of other RALMs can be found in Appendix A.}
\vspace{-0.2cm}

\label{Fig_mul_lingual_main} 
\end{figure*}

\begin{table}[t]
    \centering
    \small
    \resizebox{0.48\textwidth}{!}{
    \begin{tabular}{lcccccc}
        \toprule
        & \textbf{Prompt} & \textbf{en} & \textbf{pt} & \textbf{zh} & \textbf{ar} & \textbf{avg} \\
        \midrule
        \multicolumn{7}{c}{\textit{Flexible Language Setting}}\\

        \multirow{2}*{Qwen2-7B-Instruct} & Vanilla & \textbf{69.35} & 62.74 & 45.21 & 32.15 & 52.36 \\
         & Refined & 68.89 & \textbf{64.87} & \textbf{51.55} & \textbf{32.66} & \textbf{54.49} \\
         \cmidrule{2-7}
        \multirow{2}*{GPT-3.5-Turbo} & Vanilla & 65.14 & 61.43 & 34.86 & 32.38 & 48.46 \\
         & Refined & \textbf{73.78} & \textbf{72.56} & \textbf{50.85} & \textbf{38.99} & \textbf{59.05} \\
        \midrule
        \multicolumn{7}{c}{\textit{Strict Language Setting}} \\
        \multirow{2}*{Qwen2-7B-Instruct} & Vanilla & 27.98 &  29.03 &  20.99 & \textbf{20.60} & 24.65 \\
         & Refined & \textbf{29.04} & \textbf{29.76} & \textbf{23.30} & 19.97 & \textbf{25.52} \\
         \cmidrule{2-7}
        \multirow{2}*{GPT-3.5-Turbo} & Vanilla & 21.45 & 22.40 & 8.43 & 7.16 & 14.86 \\
         & Refined & \textbf{26.83} & \textbf{26.09} & \textbf{12.86} &\textbf{10.45} & \textbf{19.06} \\
        \bottomrule
    \end{tabular}
    }
    \caption{Comparison of performance between using the Vanilla prompt and the Refined prompt.}
    \vspace{-0.3cm}
    \label{tab:refined_prompt}
\end{table}

\begin{figure}[t] 
\centering 
\includegraphics[width=0.46\textwidth, height=0.21\textheight]{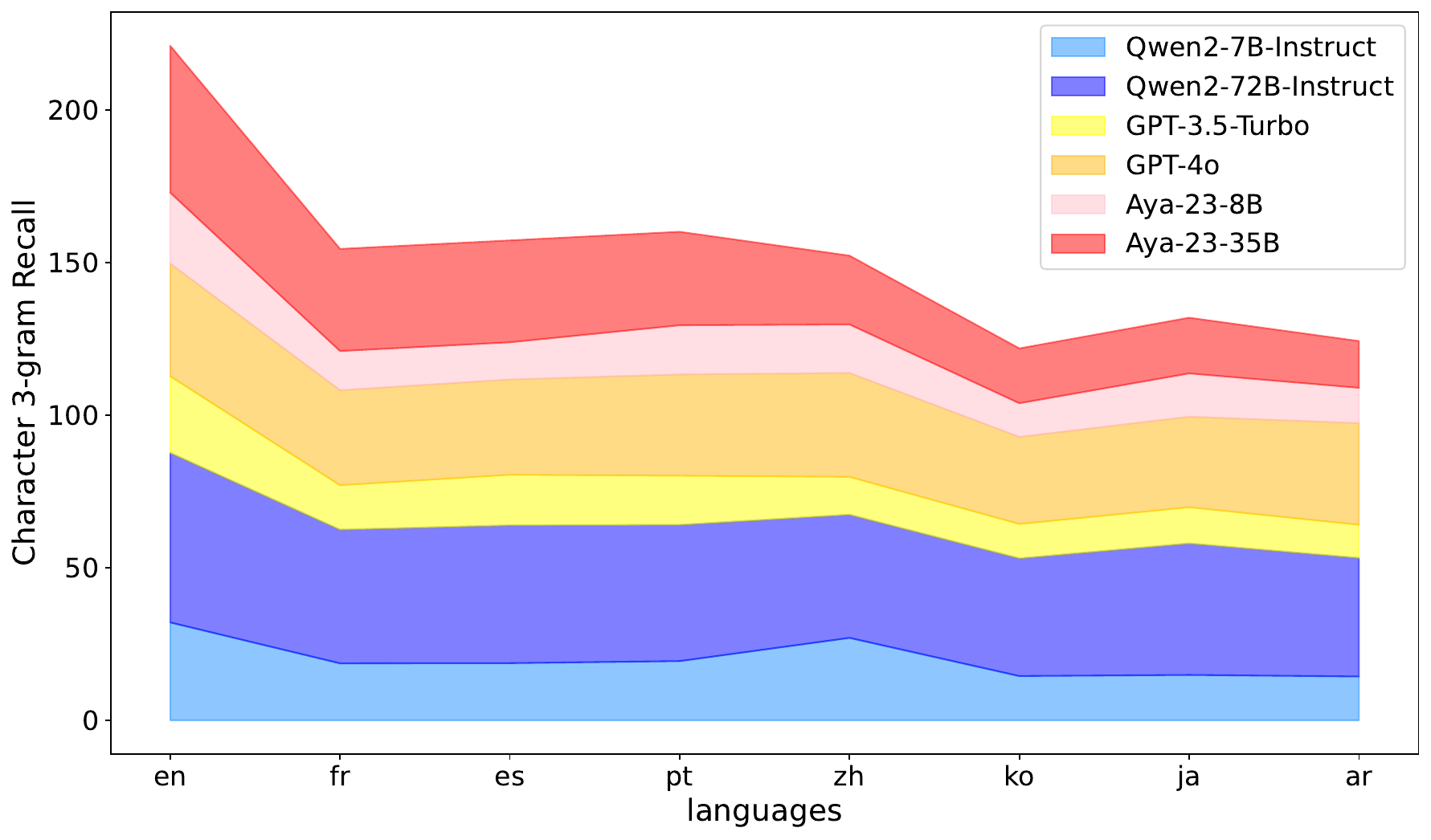} 
\caption{The stacked bar chart of Character 3-gram Recall scores for answers from different answers in the multilingual knowledge selection, with higher scores indicating a stronger preference for information from that language.} 
\vspace{-0.2cm}
\label{Fig_mul_lingual_stack} 
\end{figure}

\begin{figure}[t!] 
\centering 
\includegraphics[width=0.38\textwidth, height=0.450\textheight]{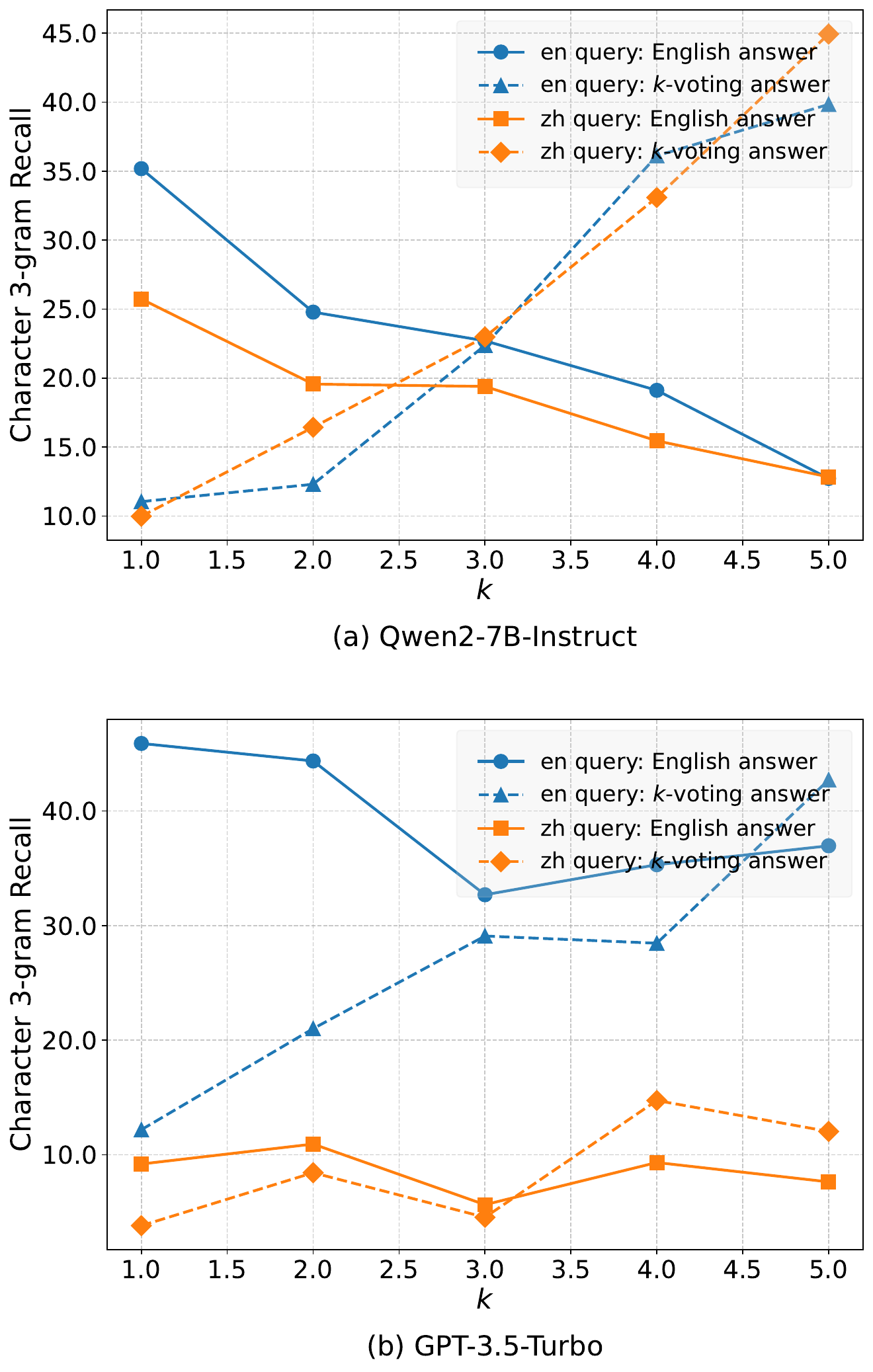} 
\caption{The impact of increasing the number of Non-English documents (denoted as $k$) on RALMs.} 
\vspace{-0.1cm}
\label{Fig_mul_voting} 
\end{figure}

\subsection{3.5\quad Experiments on Multilingual Knowledge Selection} 

This task assesses RALMs' selection bias toward languages within a knowledge conflict scenario. In this group of experiments, we present RALMs with eight randomly ordered documents in different languages, each providing a distinct answer\footnote{Since our data is fictional, we consider all these answers as potentially correct.}. By comparing the Character 3-gram Recall for answers from different languages, we can assess RALMs' language preference. The results shown in Figure \ref{Fig_mul_lingual_main} lead us to the following conclusions:

\paragraph{English benefits from selection bias and speaks louder in multilingual contexts.} To better analyze the RALMs' preference for different languages, we aggregate the Character 3-gram Recall for answers from various languages and present the cumulative results in Figure \ref{Fig_mul_lingual_stack}. Results imply that English consistently ranks as the most preferred language for providing answers, and other Indo-European languages (such as French, Spanish, and Portuguese) follow for most RALMs. An exception is Qwen2, which exhibits a greater preference for Chinese even though English remains the primary choice in most cases. We believe this is because Qwen2 has been specifically enhanced for the Chinese.

Meanwhile, we observe that RALMs favor query languages. As shown in Figure \ref{Fig_mul_lingual_main}, the values on the diagonal lines that assess RALM's preference for the query languages are typically higher. This trend is particularly pronounced for GPT-4o, whose preference for the query language can even exceed that for English. For example, when the query language is Portuguese, GPT-4o shows a significantly higher recall score for Portuguese (48.09) compared to English (35.77).

Based on the experimental results, we find that RALMs show a significant selection bias and a marked preference for English. This tendency may result in the neglect of non-English documents, which is harmful to multilingual RAG. To address this issue, we aim to explore the following question: \textit{How can we mitigate the selection bias of RALMs toward English?} Inspired by previous studies, we explore leveraging two well-known characteristics of RAG: the \textit{majority rule} and its \textit{sensitivity to document positions} to tackle this issue. As in the prior experiment, we still use Qwen2-7B-Instruct and GPT-3.5-Turbo as our RALMs.

\begin{figure}[t!] 
\centering 
\includegraphics[width=0.38\textwidth, height=0.200\textheight]{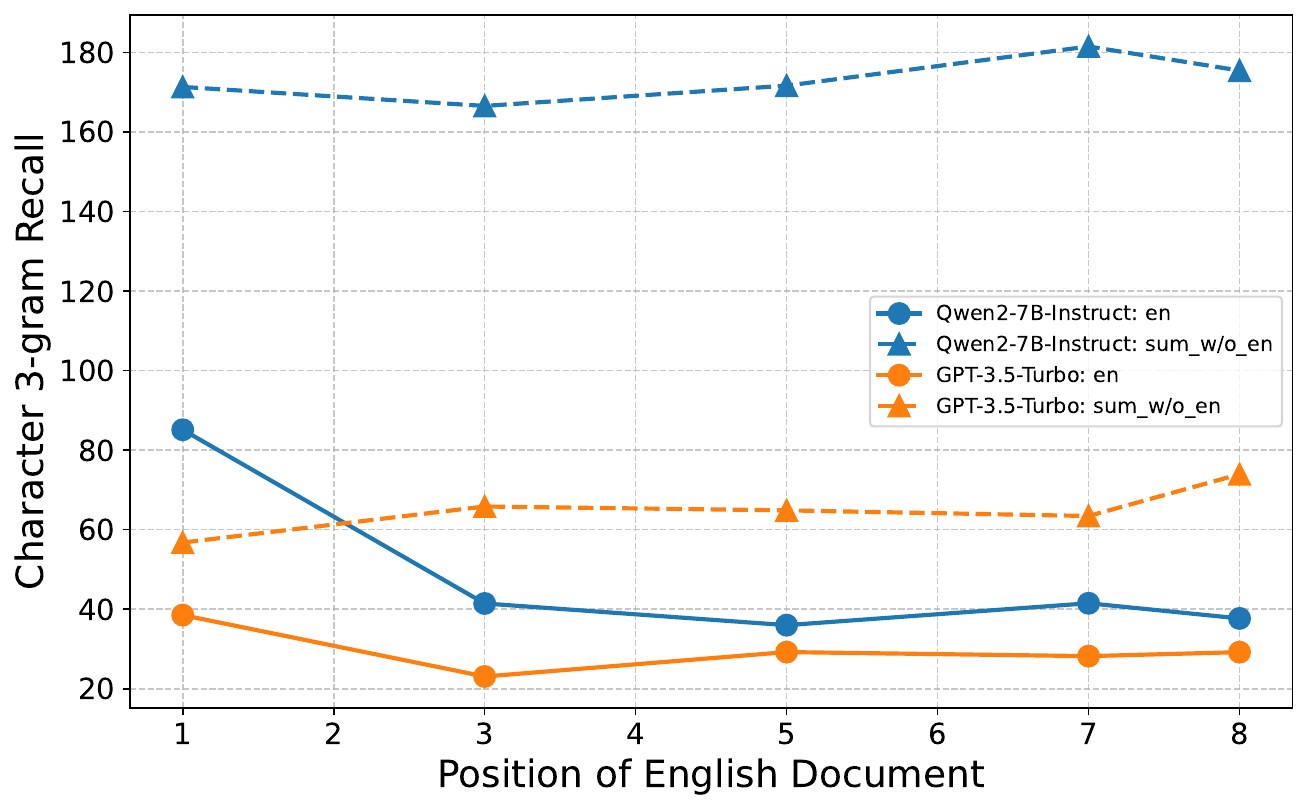} 
\caption{The impact of altering the position of English documents on RALMs.} 
\vspace{-0.4cm}
\label{Fig_mul_pos} 
\end{figure}

The majority rule suggests that RALMs favor answers that appear more frequently in the documents \cite{jin-etal-2024-tug}. Based on this characteristic, we construct an experimental setup that includes $k$ non-English documents containing the same answer, referred to \textit{$k$-voting answer}, while the remaining documents still present different answers. By comparing the Character 3-gram Recall for English answers and the \textit{$k$-voting answer}, we can assess how many non-English documents are needed to alleviate the RALMs' preference for English.

Figure \ref{Fig_mul_voting} shows the experimental results when using Chinese and English queries. The results indicate that when the $k$ is low, the Character 3-gram Recall for English answers is significantly higher than that for $k$-voting answers. However, as $k$ increases, the scores for $k$-voting answer rise while those for the English answers decline. When $k$ reaches 3, the scores for English answers and voting answers are comparable for Qwen2-7B-Instruct. We also see the same trend with GPT-3.5-Turbo when $k$ is 4. Based on these observations, we recommend introducing more non-English documents to mitigate the selection bias of RALMs toward English.

Moreover, RALMs are known for their sensitivity to document positions. \citet{liu-etal-2024-lost} reveals that the position of the gold documents within the context can significantly impact RALMs' performance. To investigate this, we alter the positions of English documents in the context and conduct experiments with English queries. From results presented in Figure \ref{Fig_mul_pos}, we can find that when English documents are placed in the first position, RALMs achieve the highest Character 3-gram Recall score for English answers (indicated by $en$), while the score for non-English answers (indicated by $sum\_w/o\_en$) is typically low. In contrast, when English documents occupy non-initial positions, the scores exhibit an opposite trend, indicating a decrease in RALMs' selection bias towards English and a heightened emphasis on non-English documents. Thus, we advise positioning English documents in non-initial positions to help mitigate RALMs' selection bias toward English.

\section{Related Work}
\paragraph{Retrieval-Augmented Generation} RAG enhances LLMs by integrating relevant texts from external knowledge resources. Most of the current RALMs focus on query refining or better document utilization. For example, Rewrite-Retrieve-Read \cite{ma2023queryrewritingretrievalaugmentedlarge} trains a small language model as the rewriter to better align the query to the retriever and the LLM reader. Chain of Note \cite{yu2023chainofnoteenhancingrobustnessretrievalaugmented} generates reading notes for retrieved documents to improve the robustness of RALMs in facing noisy, irrelevant documents and in handling unknown scenarios. However, these studies primarily focus on English, lacking exploration in multilingual RAG scenarios.

Although, many benchmarks \cite{chen2023benchmarkinglargelanguagemodels,lyu2024crudragcomprehensivechinesebenchmark,liu2023recallbenchmarkllmsrobustness,thakur2024nomiraclknowingdontknow} are proposed to evaluate RALMs' performance. Multilingual benchmarks like MKQA \cite{mkqa} and XOR-TyDi QA \cite{asai-etal-2021-xor} are time-sensitive and have potential leakage risk, while NoMIRACL\cite{thakur2024nomiraclknowingdontknow} lacks ground truth answers, making it unable to assess RALMs in comprehension and response generation.

\paragraph{Multilingualism in LLM}
Multilingual Large Language Models (MLLMs) achieve remarkable success thanks to multilingual datasets such as mC4 \cite{xue2021mt5massivelymultilingualpretrained}, CulturaX \cite{nguyen2023culturaxcleanedenormousmultilingual}, Aya Dataset \cite{singh2024ayadatasetopenaccesscollection}, and MultilingualSIFT \cite{ChenMultilingualSIFTMultilingualSupervised2023}. Along with the developments of MLLMs, related analyses have also been carried out \cite{shi2022languagemodelsmultilingualchainofthought,yuan2024vocabularysharingfacilitatesmultilingualism,yang2023bigtranslateaugmentinglargelanguage, qin2024multilinguallargelanguagemodel,xu2024surveymultilinguallargelanguage}. For instance, \citet{yuan2024vocabularysharingfacilitatesmultilingualism} analyze the multilingual capability of LLMs from a vocabulary-sharing perspective, while \citet{qin2024multilinguallargelanguagemodel} study alignment methods from parameter-tuning and parameter-frozen aspects.

Most recently, \citet{chirkova2024retrievalaugmentedgenerationmultilingualsettings} conduct multilingual RAG experiments on MKQA \cite{mkqa} and XOR-TyDi QA \cite{asai-etal-2021-xor}, discovering code-switching phenomena. \citet{sharma2024fauxpolyglotstudyinformation} create a machine-translated dataset finding that RALMs tend to prefer documents in queries' language. Compared to these studies, we conduct a comprehensive analysis based on the proposed benchmark with three evaluation tasks: monolingual knowledge extraction, cross-lingual knowledge transfer, and multilingual knowledge selection. We also offer three advice based on the observations for better multilingual RAG.

\section{Conclusion}

In this paper, we propose a new time-insensitive multilingual RAG benchmark \textit{Futurepedia} for multilingual RAG. Our benchmark includes parallel documents and corresponding QA in eight languages, where three evaluation tasks are introduced: monolingual knowledge extraction, cross-lingual knowledge transfer, and multilingual knowledge selection.
Then, we conduct experiments to evaluate the commonly used multilingual RALMs. Our findings reveal significant linguistic inequality: 1) high-resource languages stand out in the task of monolingual knowledge extraction; 2) Indo-European languages lead RALMs to provide answers directly from documents in cross-lingual knowledge transfer, alleviating the challenge of expressing answers across languages; 3) English benefits from RALMs’ selection bias and speaks louder in multilingual knowledge selection. Based on these findings, we try some strategies to improve multilingual RALMs. In the future, we will explore ways to mitigate or leverage the linguistic inequality of multilingual RAG.

\bibliography{aaai25}

\newpage

\appendix
\label{sec:appendix}

\section{Supplementary Experimental Results}
\subsection{A.1\quad Cross-Lingual Knowledge Transfer}

\begin{minipage}{\textwidth}
    \centering
    \adjustbox{max size={1.15\textwidth}{0.90\textheight}}{
        \includegraphics{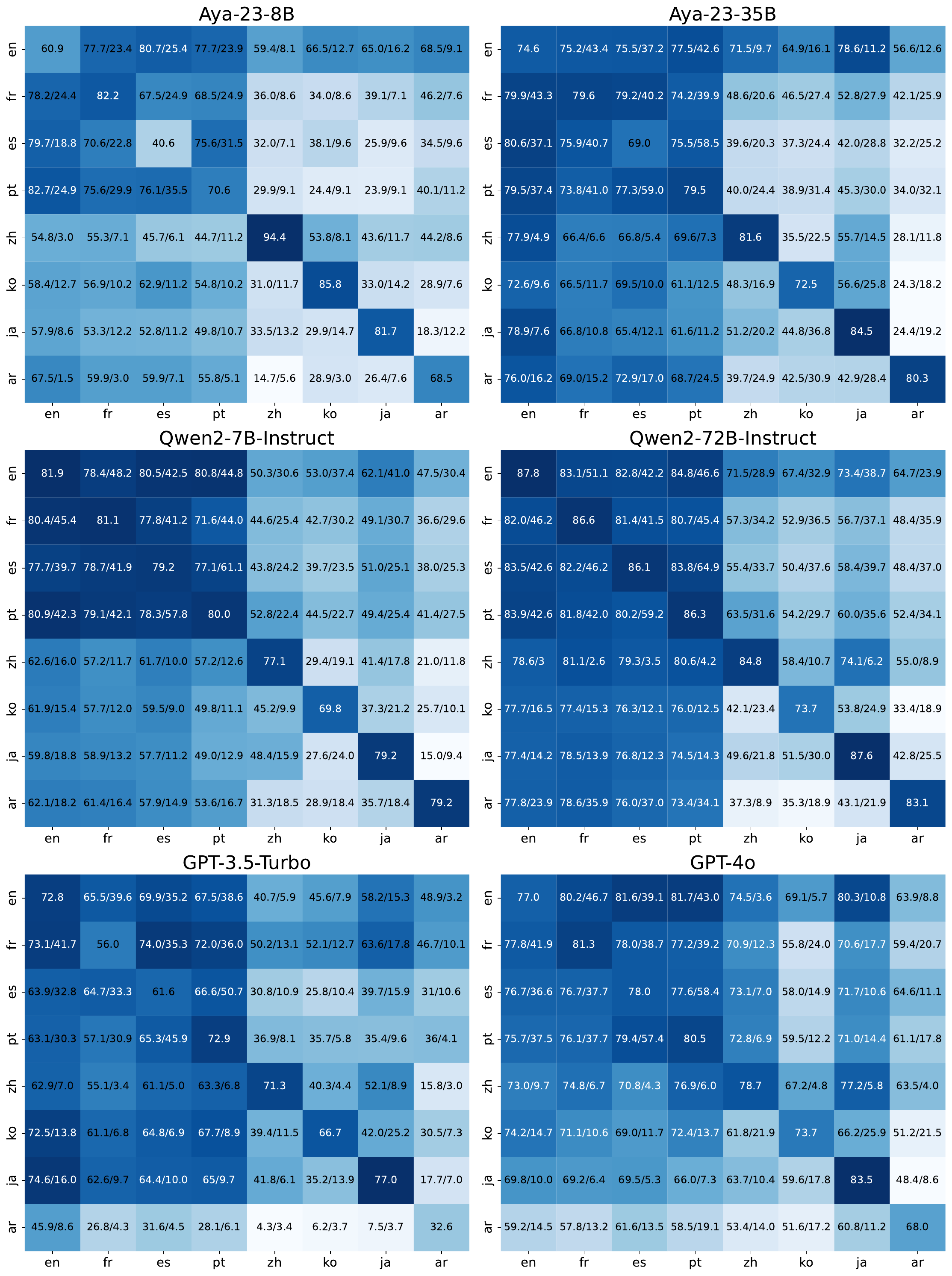}
    }
    \captionsetup{labelformat=empty}
    \captionsetup[figure]{hypcap=false}
    \captionof{figure}{ Figure 9:  Performance of RALMs on cross-lingual knowledge transfer using Character 3-gram Recall as metric. Note the first/second values represent the RALM performance in the Flexible/Strict Language Setting. The colors indicate the performance of the strict language setting, with deeper blues representing a stronger performance.}
\end{minipage}
\clearpage

\begin{figure}[p]
    \centering
    \begin{minipage}{\textwidth}
        \centering
        \vspace*{-2cm}

        \adjustbox{max size={1.15\textwidth}{0.96\textheight}}{
            \includegraphics{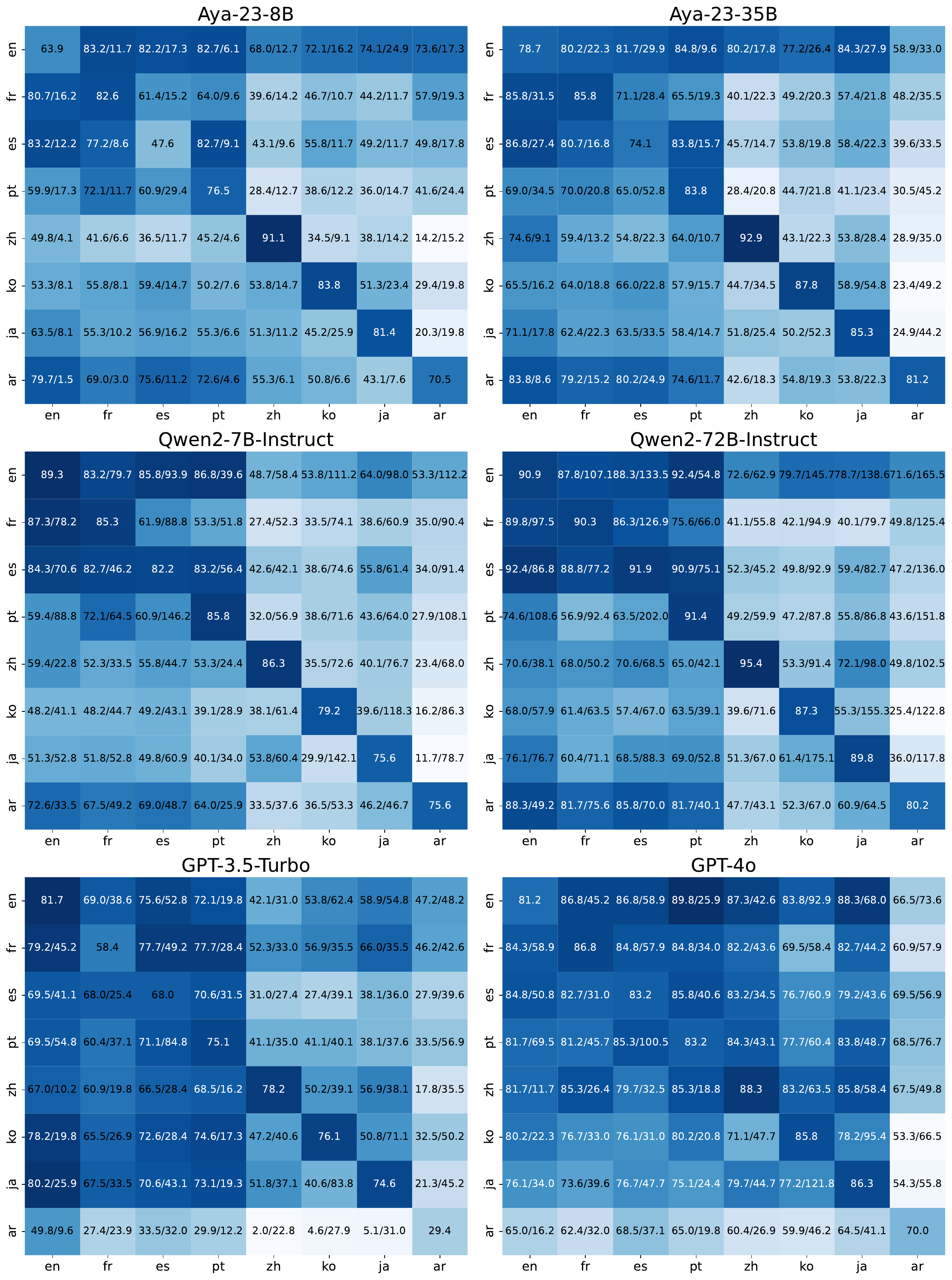}
        }
        \caption*{Figure 10: Performance of RALMs on cross-lingual knowledge transfer when using LLM for evaluation. Specifically, we employ GPT-4-Turbo to assess the predictions of RALMs.}
    \end{minipage}
\end{figure}
\clearpage

\subsection{A.2\quad Multilingual Knowledge Selection}
\begin{minipage}{\textwidth}
    \centering
    \adjustbox{max size={1.15\textwidth}{0.90\textheight}}{
        \includegraphics{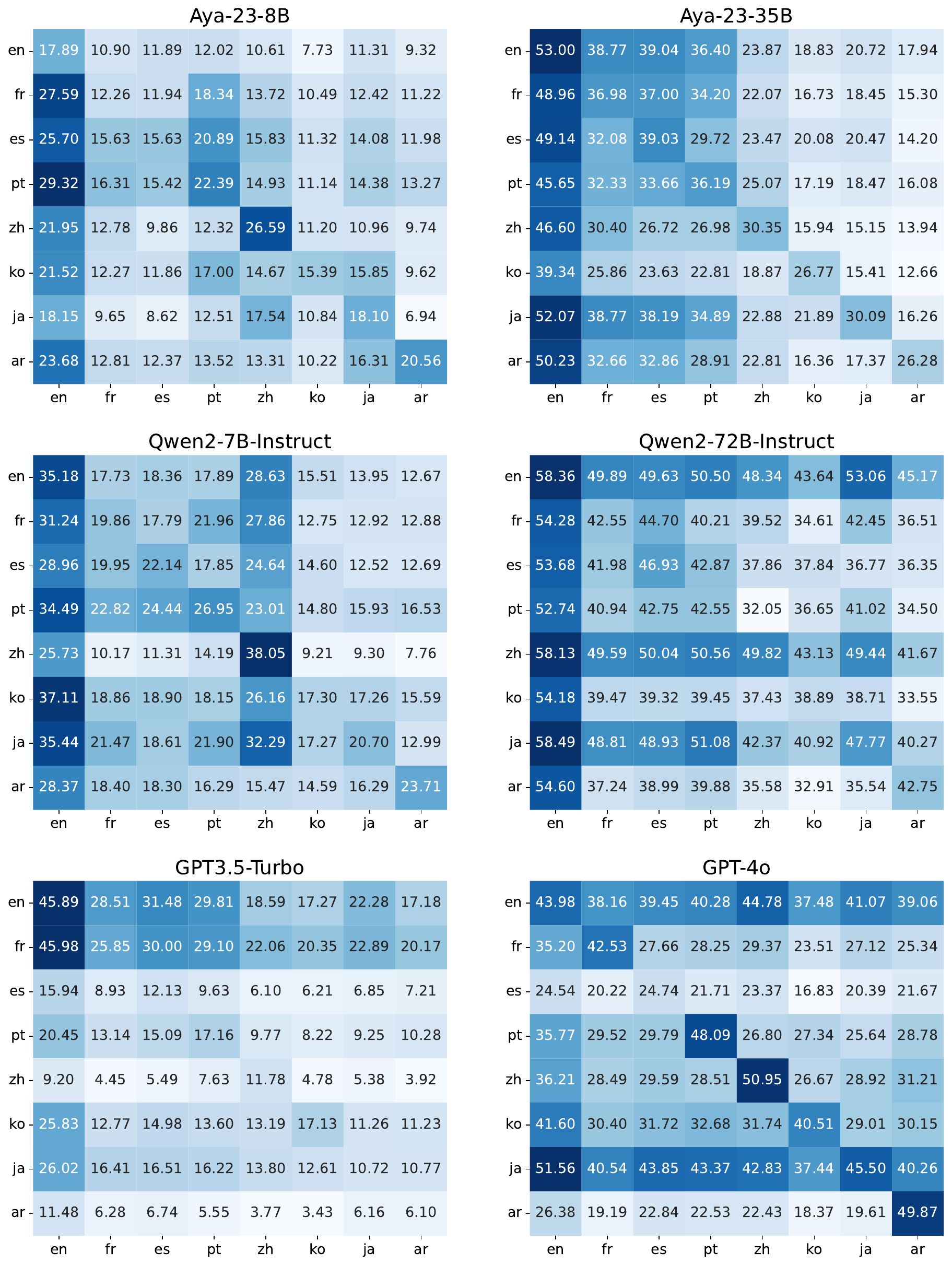}
    }
    \captionsetup{labelformat=empty}
    \captionsetup[figure]{hypcap=false}
    \captionof{figure}{Figure 11:  Performance of RALMs on multilingual knowledge selection using Character 3-gram Recall as metric. The colors indicate the performance of the strict language setting, with deeper blues representing a stronger performance.}
\end{minipage}
\clearpage

\begin{minipage}{\textwidth}
    \centering
    \adjustbox{max size={1.15\textwidth}{0.90\textheight}}{
        \includegraphics{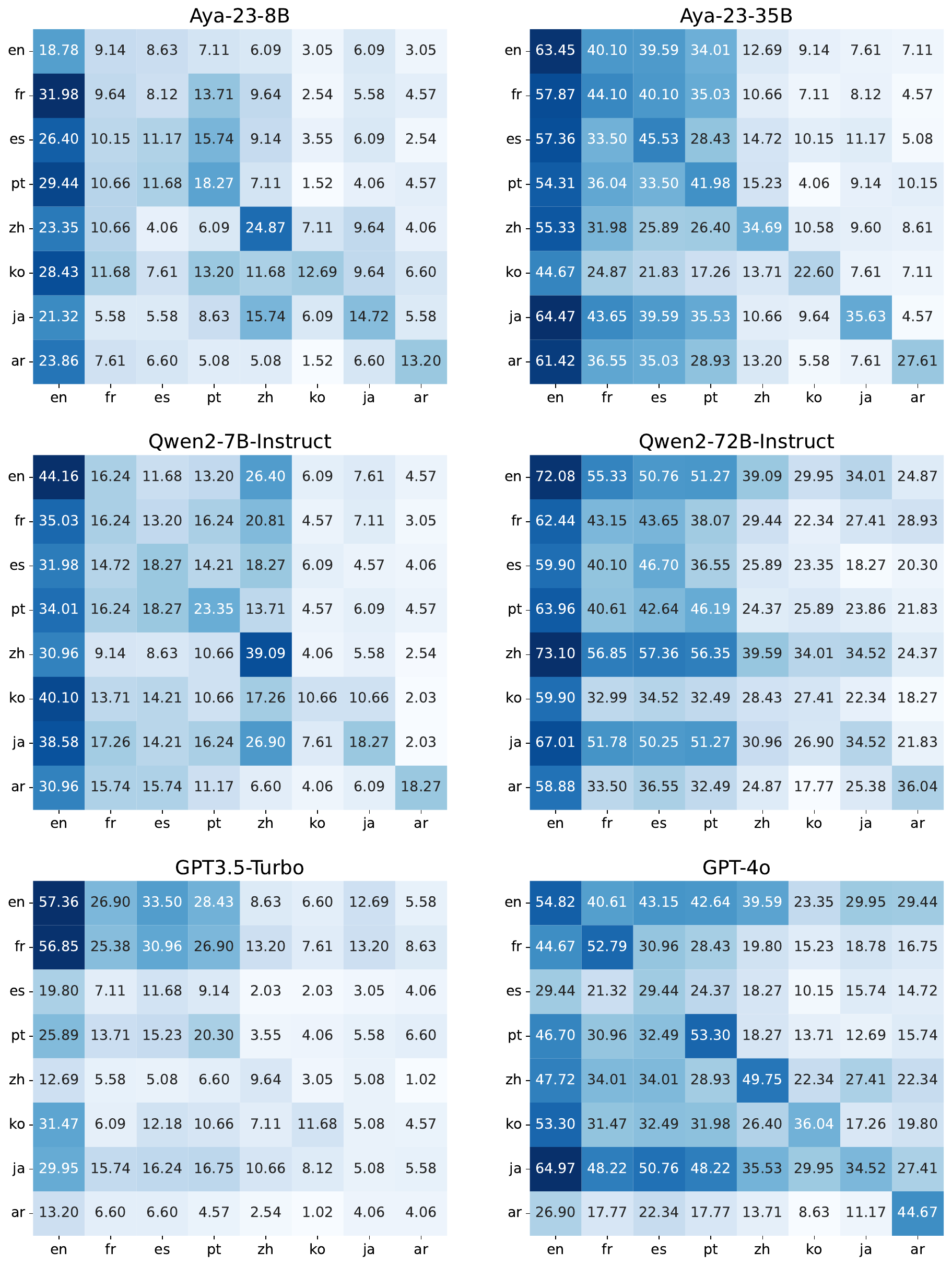}
    }
    \captionsetup{labelformat=empty}
    \captionsetup[figure]{hypcap=false}
    \captionof{figure}{Figure 12:  Performance of RALMs on multilingual knowledge selection when using LLM for evaluation. Specifically, we employ GPT-4-Turbo to assess the predictions of RALMs.}
\end{minipage}
\clearpage

\subsection{A.3\quad Experiments on the Refined Prompt}
\vspace{0.25cm}
\begin{minipage}{\textwidth}
    \centering
    \resizebox{0.88\textwidth}{!}{
    \begin{tabular}{lcccccccccc}
        \toprule
        & \textbf{Prompt} & \textbf{en} &\textbf{fr} &\textbf{es} & \textbf{pt} & \textbf{zh} & \textbf{ko} & \textbf{ja} & \textbf{ar} & \textbf{avg} \\
        \midrule

        \multicolumn{11}{c}{\textit{Flexible Language Setting}}\\

        \multirow{2}*{Qwen2-7B-Instruct} & Vanilla & \textbf{69.35} & 67.34  & 67.62 & 62.74 & 45.21 & 37.98 & 46.57 & 32.15 & 52.36 \\
         & Refined & 68.89 & \textbf{69.08} & \textbf{68.63} & \textbf{64.87} & \textbf{51.55} & \textbf{42.22} & \textbf{50.53} & \textbf{32.66} & \textbf{56.05} \\
         \cmidrule{2-11}
        \multirow{2}*{GPT-3.5-Turbo} & Vanilla & 65.14 & 56.13 & 61.58 & 61.43 & 34.86 & 34.43 & 42.67 & 32.38 & 48.58 \\
         & Refined & \textbf{73.78} & \textbf{64.54} & \textbf{71.68} & \textbf{72.56} & \textbf{50.85} & \textbf{44.77} & \textbf{55.67} & \textbf{38.99} & \textbf{59.11} \\
        \midrule

        \multicolumn{11}{c}{\textit{Strict Language Setting}} \\
        \multirow{2}*{Qwen2-7B-Instruct} & Vanilla & 27.98 & 26.51 & \textbf{26.64} &  29.03 &  20.99  & 25.05 & \textbf{25.65} & \textbf{20.60} & 25.31 \\
         & Refined & \textbf{29.04} & \textbf{26.76} & 26.47 & \textbf{29.76} & \textbf{23.30} & \textbf{25.91} & 25.52 & 19.97 & \textbf{25.84} \\
         \cmidrule{2-11}
        \multirow{2}*{GPT-3.5-Turbo} & Vanilla & 21.45 & 18.30 & 20.40 & 22.40 & 8.43 & 8.40 & 13.78 & 7.16 & 15.04 \\
         & Refined & \textbf{26.83} & \textbf{20.10} & \textbf{26.17} & \textbf{26.09} & \textbf{12.86} & \textbf{13.85} & \textbf{20.31} &\textbf{10.45} & \textbf{19.58} \\
        \bottomrule
    \end{tabular}
    }
    \captionof{table}{Comparison of performance between using the
Vanilla prompt and the Refined prompt across all languages.}
    \vspace{0.3cm}
    \label{tab:refined_all}
\end{minipage}
As mentioned in \S 3.4, we refine the original prompt, requiring RALMs to first provide answers from documents in various languages, and then translate them into the query language. This method leverages the chain-of-thought to enhance RALMs' performance in cross-lingual knowledge transfer. Table \ref{tab:refined_all} presents the experimental results of Qwen2-7B-Instruct and GPT-3.5-Turbo across all languages, confirming the effectiveness of our method.

\section{Implementation Details}
When collecting data, we utilized MediaWiki Action API\footnote{\url{https://www.mediawiki.org/wiki/API:Main_page}} to gather Wikipedia data. During evaluation, we simplify the retrieval process by directly providing the gold documents. Each document is divided into 200-token chunks, and we employ mcontriever-msmarco\footnote{\url{https://huggingface.co/facebook/mcontriever-msmarco}} as the retriever to select the 5 most relevant chunks for the QA as context for length control. 
We limit the RALMs' context input to 8,192 tokens and instruct RALMs that the current year is 2200 to ensure that the timestamps in the QA pairs are reasonable. RALMs are also instructed to respond based on the provided context or reply that the information is insufficient.

\section{Instructions in the Experiments}
In this section, we provide the English instructions used for the data construction and experiments. Instructions in other languages were derived through translation from the English version.

\newpage

\textcolor{white}{ }\\
\textcolor{white}{ }\\
\textcolor{white}{ }\\
\textcolor{white}{ }\\
\textcolor{white}{ }\\
\textcolor{white}{ }\\
\textcolor{white}{ }\\
\textcolor{white}{ }\\
\textcolor{white}{ }\\
\textcolor{white}{ }\\
\textcolor{white}{ }\\
\textcolor{white}{ }\\
\textcolor{white}{ }\\
\textcolor{white}{ }\\
\textcolor{white}{ }\\
\textcolor{white}{ }\\
\textcolor{white}{ }\\

\textbf{Instruction for Entities Modification}\\
\bigskip
\noindent\fbox{
    \parbox{0.97\linewidth}{Given the original entity \{ENTITY\}, you should provide 8 unique and reasonable entities of the same type as the original entity. If the given entity is a person's name, a movie title, a game title, etc., return a fictional but reasonable entity. If the given entity is a country or a city, provide a real and comparatively similar entity. Return the result in the form of Python lists, with no additional context.
    }
}

\textbf{Instruction for Document Update}\\
\bigskip
\noindent\fbox{
    \parbox{0.97\linewidth}{Your task is to modify a given document by replacing certain entities within it to ensure logical coherence and consistency. Please adhere to the following instructions:\\
1. The modified document must be relevant and capable of answering the question: \{QUERY\} with the answer \{ANSWER\}.\\
2. Ensure all mentioned entities are thoroughly replaced. For example, if replacing a person's name, ensure both the first and the last name are replaced everywhere.
3. For entities with aliases (e.g., “United States” and “America” in English), replace all variations.
4. After replacing an entity (e.g., changing “USA” to “UK”), adjust related content accordingly (e.g., changing “American” to “British”).
    }
}

\textbf{Instruction for Monolingual Knowledge Extraction}
\noindent\fbox{
    \parbox{0.97\linewidth}{
        You are an accurate and reliable AI assistant that can answer queries with the help of external documents, and you need to use the same language as the query to give your answer. If the information in the documents does not contain the answer, you will generate ``The document contains insufficient information, so I cannot answer the query based on the document.'' If the information in the documents contains the correct answer, you will succinctly and directly give all the answers without including any context, and if there are multiple answers, separate them by ``, ''. Note that the current time is the year 2200, and the temporal information, names, and other entity information mentioned in the text are all correct. Now the Document is :\{DOCS\} ... the query is:\{QUERY\}"
    }
}
\newpage
\textbf{Instruction for Cross-lingual Knowledge Transfer}\\
\bigskip
\noindent\fbox{
    \parbox{0.97\linewidth}{
        You are an accurate and reliable AI assistant that can answer queries with the help of external documents in different languages, and you need to use the same language as the query to give your answer. If the information in the documents does not contain the answer, you will generate ``The document contains insufficient information, so I cannot answer the query based on the document.'' If the information in the documents contains the correct answer, you will succinctly and directly give all the answers without including any context, and if there are multiple answers, separate them by ``, ''. Note that the current time is the year 2200, and the temporal information, names, and other entity information mentioned in the text are all correct. Now the Document is :\{DOCS\} ... the query is:\{QUERY\}"
    }
}
\vspace{0.1em}

\bigskip
\textbf{Refined Instruction for Cross-lingual Knowledge Transfer}\\
\bigskip
\bigskip
\noindent\fbox{
    \parbox{0.97\linewidth}{
        You are an accurate and reliable AI assistant that can answer queries with the help of external documents in different languages, and you need to use the same language as the query to give your answer. If the information in the documents does not contain the answer, you will generate ``The document contains insufficient information, so I cannot answer the query based on the document.'' If the information in the documents contains the correct answer, you will succinctly and directly give all the answers without including any context, and if there are multiple answers, separate them by ``, ''. Note you need to provide the answer from the original text and then transfer it into the query language. The format is in the format [answer in query language (answer in the original text), ...]. Note that the current time is the year 2200, and the temporal information, names, and other entity information mentioned in the text are all correct. Now the Document is :\{DOCS\} ... the query is:\{QUERY\}"
    }
}

\bigskip
\textbf{Instruction for Multilingual Knowledge Selection}
\vspace{0.3em}
\noindent\fbox{
    \parbox{0.97\linewidth}{
        You are an accurate and reliable AI assistant that can answer queries with the help of multiple external documents in various languages with different answers, and you need to use the same language as the query to give your answer. If the information in the documents does not contain the answer, you will generate ``The document contains insufficient information, so I cannot answer the query based on the document.'' Otherwise, you will succinctly and directly give all the answers you think are correct without including any context, and if there are multiple answers, separate them by ``, ''. that the current time is the year 2200, and the temporal information, names, and other entity information mentioned in the text are all correct. Now the Document is :\{DOCS\} ... the query is:\{QUERY\}"
    }
}
\vspace{0.1em}
\newpage
\section{Question Construction Guidelines}
Below are the annotation guidelines for manually reviewing and modifying data to ensure quality.

\noindent\fbox{
    \parbox{0.97\linewidth}{
        This task requires reviewing the provided documents and QA to ensure they meet the following requirements; if they do not, modifications must be made:
\begin{itemize}
    \item The document must contain the knowledge mentioned in the QA and be able to answer the QA.
    \item The document must not contain conflicting information.
    \item If the QA in the target language differs from the English QA, modify it according to the English version.
    \item Ensure that the timestamps in the document and QA are consistent, with all dates falling between the years 2124 and 2200.
\end{itemize}
    How to modify the QA or the document:
    \begin{itemize}
    \item If there is a significant difference between the QA and the English QA, a re-translation is required.
    \item If the document lacks information to answer the QA, add or revise sentences to include the necessary details.
    \item Ensure all instances of names, last names, and aliases are thoroughly consistent. If the main character is named Michael Johnson, it is essential to maintain consistency throughout. Use pronouns such as ``he'' (e.g., ``He went to the store''), refer to him by his last name ``Johnson'' (e.g., ``Johnson attended the meeting''), or use his full name.
    \item Resolve any conflicts present in the document.
    \begin{itemize}
    \item Adjust the document to reflect a future time without logical inconsistencies. For instance, if the document states that a film was shot in 2021 but is set to be released in 2038, the filming date should be changed to 2038 to maintain consistency.
    \item Geographical Conflicts: These typically involve discrepancies between city and country names. For example, ``New York City in Italy.''
    \end{itemize}
    \end{itemize}
    }
}

\end{document}